\documentclass[
]{elsarticle}

\usepackage{lineno,hyperref}
\modulolinenumbers[5]

\journal{Pattern Recognition}

\usepackage[rgb]{xcolor}
\usepackage{multirow}

\usepackage{graphicx}
\usepackage{amsmath}
\usepackage{amssymb}
\usepackage{dsfont}
\usepackage{booktabs}
\usepackage{eucal}
\usepackage{color, colortbl}
\usepackage{multirow}
\usepackage{tikz,pgf,pgfplots,pgfplotstable}
\usepackage{pgfmath}
\usepgfplotslibrary{statistics}
\usepgfplotslibrary{groupplots}
\usetikzlibrary{plotmarks}
\usetikzlibrary{shapes,arrows}
\usepackage{pgfmath}
\pgfplotsset{compat=1.15} 

\definecolor{Gray}{gray}{0.90}
\definecolor{Red}{rgb}{0.77,0.0,0.0}

\usepackage[]{hyperref}

\begin{document}
\definecolor{our_blue}{rgb}{.2,.6,.8}
\definecolor{our_orange}{rgb}{1,.5,.05}
\definecolor{our_green}{rgb}{.17,.63,.17}
\definecolor{our_darkblue}{rgb}{0,0,.55}
\definecolor{our_red}{rgb}{.84,.15,.16}

\pgfplotsset{
    style_line/.style={
        line width = 2pt,
    }
}

\pgfplotsset{
    style_kpconv/.style={
        style_line,
        color = our_darkblue,
        mark = diamond,
        mark size = 3.5pt,
    },
    style_convpoint/.style={
        style_line,
        color = our_blue,
        mark = o,
        mark size = 3.5pt,
    },
    style_linear/.style={
        style_line,
        color = our_orange,
        mark = square,
        mark size = 3.5pt,
    },
    style_aggr/.style={
        style_line,
        color = our_green,
        mark = triangle,
        mark size = 3.5pt,
    }
}
\begin{frontmatter}
%
\title{Composite Convolution: a Flexible Operator for Deep Learning on 3D Point Clouds}

\author[mymainaddress]{Alberto Floris}
\author[mymainaddress]{Luca Frittoli\corref{mycorrespondingauthor}}
\cortext[mycorrespondingauthor]{corresponding author}
\ead{luca.frittoli@polimi.it}

\author[mysecondaryaddress]{Diego Carrera}
\author[mymainaddress]{Giacomo Boracchi}

\address[mymainaddress]{DEIB, Politecnico di Milano, via Ponzio 34/5, Milan (Italy)}
\address[mysecondaryaddress]{STMicroelectronics, via Camillo Olivetti 2, Agrate Brianza (Italy)}

\begin{abstract}
Deep neural networks require specific layers to process point clouds, as the scattered and irregular location of 3D points prevents the use of conventional convolutional filters. We introduce the composite layer, a flexible and general alternative to the existing convolutional operators that process 3D point clouds. We design our composite layer to extract and compress the spatial information from the 3D coordinates of points and then combine this with the feature vectors. Compared to mainstream point-convolutional layers such as ConvPoint and KPConv, our composite layer guarantees greater flexibility in network design and provides an additional form of regularization. To demonstrate the generality of our composite layers, we define both a convolutional composite layer and an aggregate version that combines spatial information and features in a nonlinear manner, and we use these layers to implement CompositeNets. Our experiments on synthetic and real-world datasets show that, in both classification, segmentation, and anomaly detection, our CompositeNets outperform ConvPoint, which uses the same sequential architecture, and achieve similar results as KPConv, which has a deeper, residual architecture. Moreover, our CompositeNets achieve state-of-the-art performance in anomaly detection on point clouds. Our code is publicly available at \textcolor{blue}{\url{https://github.com/sirolf-otrebla/CompositeNet}}. 
\end{abstract}



\end{frontmatter}

%

%
%
%
%

\section{Introduction}
\label{sec:intro}
Point clouds provide a compact yet detailed representation of 3D objects and, for this reason, they are widely employed in several applications such as autonomous driving~\cite{qian20223d}, topography~\cite{ao2017one}, architecture~\cite{su2022dla}, and industrial quality control~\cite{frittoli2022deep}. 
Point clouds are unordered sets of points in $\mathbb{R}^3$, typically acquired by LiDAR sensors~\cite{ao2017one, li2023deepsir}, Mobile Laser Scanners~\cite{su2022dla}, or depth/stereo cameras~\cite{dai2017scannet,armeni20163d}. Often, each point is paired with a feature vector such as RGB components or the normal vector to the surface of the object at that point~\cite{QiPointnet}.

The design of deep neural networks that can process point clouds has been attracting more and more interest in 3D deep learning~\cite{guo2020deep}. Such flourishing literature is motivated by the intrinsic challenges of training machine-learning models on point clouds. In fact, unlike images, point clouds are sets and not arranged on a regular grid, where most neural network layers are defined. For this reason, several \emph{point-convolutional} layers, namely layers implementing convolutions on point clouds, have been proposed in the literature, the two most representative examples being \emph{KPConv}~\cite{HuguesKPConv} and \emph{ConvPoint}~\cite{BoulchConvpoint}. 


\begin{figure}[t]
    \centering
    \usetikzlibrary{shapes,arrows}

\usetikzlibrary{positioning}
\usetikzlibrary{3d} 

\def\ConvColor{rgb:yellow,5;red,2.5;white,5}
\def\ConvReluColor{rgb:yellow,5;red,5;white,5}
\def\PoolColor{rgb:red,1;black,0.3}
\def\UnpoolColor{rgb:blue,2;green,1;black,0.3}
\def\FcColor{rgb:blue,5;red,2.5;white,5}
\def\FcReluColor{rgb:blue,5;red,5;white,4}
\def\SoftmaxColor{rgb:magenta,5;black,7}   
\def\SumColor{rgb:blue,5;green,15}

\newcommand{\copymidarrow}{\tikz \draw[-Stealth,line width=0.8mm,draw={rgb:blue,4;red,1;green,1;black,3}] (-0.3,0) -- ++(0.3,0);}

\begin{tikzpicture}

\begin{axis}[
    width=0.5\textwidth,
    height=0.4\textwidth,
    scale only axis,
    xmin=-2.75, xmax=5.25,
    ymin=-3.5, ymax=3,
    axis line style={draw=none},
    tick style={draw=none},
    xticklabels={,,},
    yticklabels={,,}
    ]

\addplot[only marks, color=our_red, mark=diamond*, mark size=3pt]
coordinates{
(0,0)
(4,0)
(2.8,1.2)
(3.5,-0.5)
(4.7,-0.8)
};

\filldraw[fill=none, thick](0,0) circle (1.2);
\filldraw[fill=none, dotted](-1.2,1.2) circle (1.2);
\filldraw[fill=none, dotted](-0.5,-0.5) circle (1.2);
\filldraw[fill=none, dotted](0.5,-0.8) circle (1.2);

\addplot[only marks, color=our_blue, mark=diamond*, mark size=3pt]
coordinates{
(-1.2,1.2)
(-0.5,-0.5)
(0.5,-0.8)
(-0.4,2)
(-0.4,0.8)
(-2,1)
(-1.35,-0.7)
(-0.1,-1.4)
(0.4,1)
(1.2,-1.)
(-1.5,2.2)
};

\node[below] at (0.1,-0.05) {\scriptsize$y$};
\node[below] at (4.1,-0.05) {\scriptsize$y$};
\node[] at (1,1) {\scriptsize$X_y$};
\node[] at (0,2.7) {\scriptsize$P$};
\node[] at (4,2.7) {\scriptsize$Q$};
\draw[->, ultra thick] (1.5,0)--(2.5,0);

\node[] at (-1.1,-2.5) {\scriptsize$X_y$};
\node[] at (-1.1,-3) {\scriptsize$\phi$};
\draw[->] (-.65,-2.5)--(-.2,-2.5);
\filldraw[fill=\SumColor, fill opacity=0.5] (-0.1,-2.76) rectangle (1.3,-2.24);
\node[] at (.6,-2.5) {\scriptsize$s(\cdot -y)$};
\draw[->] (1.4,-2.5)--(1.8,-2.5);
\draw[->] (-.65,-3)--(1.8,-3);
\filldraw[fill=\ConvColor, fill opacity=0.5] (1.9,-3.25) rectangle (3.1,-2.25);
\node[] at (2.5,-2.75) {\scriptsize$f$};
\draw[->] (3.2,-2.75)--(3.75,-2.75);
\node[right] at (3.75,-2.75) {\scriptsize$\psi$};

\end{axis}

\end{tikzpicture}
    \vspace{-0.75cm}
    \caption{The operations of our composite layer on the input point cloud $P$ (the blue dots $x$, each paired with a feature vector $\phi(x)$) to obtain the output point cloud $Q$ (the red dots $y$ sampled from $P$, each paired with its output feature vector $\psi(y)$). The spatial function $s$ outputs a vector in $\mathbb{R}^K$ for each point $x$ belonging to the convolution window $X_y$, where $y$ is the output point. The semantic function combines the input features $\phi$ and the output of the spatial function $\{s(x-y)\}_{x \in X_y}$ to produce the output features $\psi(y)$. 
    }
    \label{fig:composite-general}
    \vspace{-0.3cm}
\end{figure}

In this work 
we introduce the \emph{composite layer} (Figure~\ref{fig:composite-general}), a flexible and general alternative to the existing point-convolutional layers in deep neural networks for point clouds. We define the composite layer as the composition of two learnable functions defined over a \emph{convolution window} $X_y$, a neighborhood of each point $y$ of the output point cloud. The first learnable function, called \emph{spatial function} $s$, extracts and compresses the information from the spatial arrangement of the points in $X_y$. The second, called \emph{semantic function} $f$, combines the spatial information extracted by $s$ with the feature vectors $\phi(x)\in\mathbb{R}^I$ associated with each point $x \in X_y$. Similarly to convolutional layers for images, our semantic function stacks $J$ \emph{filters} taking as input the spatial information extracted by $s$ and all the input features to produce the output features $\psi(y)\in\mathbb{R}^J$. 

The most relevant advantage of composite layers is that the spatial and semantic functions can be customized independently, guaranteeing a general formulation and remarkable flexibility when tuning the network hyperparameters. In contrast, existing point-convolutional layers such as the mainstream KPConv~\cite{HuguesKPConv}, ConvPoint~\cite{BoulchConvpoint}, and their more recent variants~~\cite{yang2021adaptive, mao2019interpolated, lin2020fpconv, liu2019densepoint, liu2019relation, ran2021learning} lack such flexibility and cannot be easily customized. For instance, by tuning the hyperparameters of a traditional point-convolutional layer~\cite{HuguesKPConv, BoulchConvpoint} we can improve the network performance at the cost of substantially increasing the number of parameters. In our composite layer, the same improvement can be achieved by tuning only the hyperparameters of the spatial function, which results in a negligible increase in the overall number of parameters of the network, as we empirically demonstrate in Section \ref{subsec:flexibility}. 

There are two other implications stemming from the structure of our composite layer:
\begin{itemize}
    \item Regularization: in our composite layer, the spatial and semantic functions are defined by two separate weight tensors, which are then multiplied to obtain a unique weight tensor whose rank is constrained to be lower or equal to that of the smaller factor. This low-rank property can be considered a peculiar form of regularization.
    \item Modularity: we can design new layers by simply redefining the semantic or the spatial function. To demonstrate this aspect of generality, we define two versions of the composite layer: besides the \emph{convolutional composite layer}, which is an alternative formulation of point convolution, we present the \emph{aggregate composite layer}, which combines spatial information and features in a nonlinear manner, while most existing layers are linear.
\end{itemize}





We use our composite layers to implement \emph{CompositeNets}, deep neural networks that we successfully train for classification, semantic segmentation, and unsupervised anomaly detection, following the self-supervised approach presented in~\cite{golan2018deep}. The unsupervised anomaly detection experiment tackles a rarely investigated task in 3D point cloud domain, and has been addressed only in very specific scenarios~\cite{ao2017one, lyu2021online}. 

In our experiments, we compare the classification, segmentation, and anomaly-detection performance of our CompositeNets with ConvPoint~\cite{BoulchConvpoint} and KPConv~\cite{HuguesKPConv} on the synthetic \emph{ModelNet40}~\cite{WuModelnet} and \emph{ShapeNet}~\cite{ChangShapenet} datasets, and the real-world \emph{ScanNet}~\cite{dai2017scannet} and \emph{S3DIS}~\cite{armeni20163d} datasets. In the considered tasks, our convolutional CompositeNet is on par with ConvPoint, which has the same sequential architecture, and our aggregate CompositeNet achieves similar performance as KPConv~\cite{HuguesKPConv}, which instead has a deeper, residual architecture. Hence, our CompositeNets are at least on par with existing point-convolutional networks while having superior design flexibility, as we demonstrate in Section \ref{subsec:flexibility}. Moreover, our self-supervised CompositeNets outperform both anomaly detectors based on hand-crafted descriptors~\cite{hana2018comprehensive} and the autoencoder that is the only deep anomaly-detector for point clouds in the literature~\cite{masuda2021toward}.

To summarize, our key contributions are:
\begin{itemize}
    \item We introduce the composite layer, a new layer for point-cloud processing in neural networks that is more flexible and general than the existing ones.
    \item To show their generality, we implement convolutional and aggregate (nonlinear) composite layers to be used as building blocks for CompositeNets.
    \item Our self-supervised CompositeNets achieve state-of-the-art anomaly detection performance on synthetic and real-world point cloud benchmarks.
\end{itemize}

The rest of the paper is organized as follows: in Section~\ref{sec:related_works} we review the literature on deep learning for point clouds, with a focus on point convolutions. In Section \ref{sec:problem} we give a formal description of point-convolutional operators, while in Section \ref{sec:pointconv} we provide the background on how point convolution has been implemented in the literature. Then, in Section \ref{sec:approach} we present our convolutional and aggregate composite layers. In Section \ref{sec:experiments} we illustrate our experiments, and we empirically demonstrate the superior design flexibility of our composite layers compared to existing point-convolutional layers in our ablation studies. Finally, in Section \ref{sec:conclusions} we give some conclusions and future research directions.

\section{Related Work}
\label{sec:related_works}
The first deep-learning methods for point clouds cope with their scattered nature by projecting the 3D points on several planes~\cite{liu2022vfmvac}, thus generating 2D images that can be processed by traditional CNNs. Another approach maps 3D point clouds to a voxel grid so that they can be processed by 3D convolutional filters~\cite{choy20194d}. However, both approaches imply a loss of information due to the projection or quantization of the coordinates of the points. Moreover, the computational cost of the second approach scales very poorly with the grid resolution. \emph{Submanifold sparse convolutional networks}~\cite{graham20183d} represent a viable alternative to efficiently handle higher-resolution voxel grids by leveraging the sparsity of point clouds. In what follows we overview deep-learning techniques operating directly on 3D point clouds without projections or voxelization, with a focus on point-convolutional layers. We refer to \cite{guo2020deep} for a more comprehensive review of deep learning for 3D point clouds. 



\noindent\textbf{Early approaches.} \emph{PointNet}~\cite{QiPointnet} is the first deep neural network directly processing 3D point clouds using Multi-Layer Perceptrons (MLPs) and permutation-invariant pooling. The major drawback is that PointNet feeds the entire point cloud to the same MLP and thus might fail at recognizing local structures inside the point cloud. An improved solution is \emph{PointNet++}~\cite{QiPointnet++}, which hierarchically applies PointNet to nested partitions of the point cloud. Similarly, \emph{ShellNet}~\cite{zhang2019shellnet} uses MLPs with permutation-invariant pooling to extract features from the points contained in concentric spherical neighborhoods, and then applies 1D convolutional layers.

\noindent\textbf{Point Convolutions.} Due to the success of CNNs in visual recognition problems, recent works aim at defining neural network layers for point clouds operating similarly to convolutional filters on images. In this direction, \emph{volumetric convolutions}~\cite{AtzmonPCNNEO} applies standard convolutions on 3D functions and then evaluates the resulting functions on the input point cloud. Other methods approximate continuous 3D convolutional kernels by MLPs~\cite{wang2018deep, wu2019pointconv} or polynomials~\cite{xu2018spidercnn}.

The most popular and effective solutions involve \emph{point-convolutional layers}, which process point clouds by feeding the coordinates of the points to either a \emph{Radial Basis Function Network} (RBFN) or a \emph{Multi-Layer Perceptron} (MLP) and then linearly combine the result with the input features to obtain the output features, similarly to convolutional filters on images. Here we describe two prominent examples of these layers, namely \emph{KPConv}~\cite{HuguesKPConv} and \emph{ConvPoint}~\cite{BoulchConvpoint}, and some more recent methods following their approaches~\cite{yang2021adaptive, mao2019interpolated, lin2020fpconv, liu2019densepoint, liu2019relation, ran2021learning}. 

The most representative solution following the first approach is KPConv~\cite{HuguesKPConv}, which uses an RBFN to process the coordinates of the points in a window $X_y$ (relatively to the position of the output point $y$) before combining them with the input features. KPConv and its extension \emph{AGMMConv}~\cite{yang2021adaptive} also implement a deformable convolutional kernel that adaptively defines the RBFN depending on the output point where the kernel is applied. \emph{InterpCNN}~\cite{mao2019interpolated} is very similar to KPConv, the only difference being the chosen RBFs. \emph{FPConv}~\cite{lin2020fpconv} builds on the assumption that point clouds represent locally flat surfaces and thus estimates the tangent plane to the surface at each point and projects the neighboring points to the plane. Then, FPConv processes the projected points by a 2D RBFN whose centers are positioned on a regular grid, similarly to KPConv. 

The most representative layer defined by MLPs is \emph{ConvPoint}~\cite{BoulchConvpoint}. This operator feeds an MLP with the relative coordinates of the points with respect to a set of centers, which play a similar role as the RBFN centers in KPConv. Then, the MLP outputs are linearly combined with the input features. This approach has been followed by other layers such as \emph{DensePoint}~\cite{liu2019densepoint}, \emph{RS-CNN}~\cite{liu2019relation}, and \emph{RPNet}~\cite{ran2021learning}, which however are not convolutional since the input of their MLPs includes the absolute position of the points. All the other point-convolutional layers consider the relative position of the points with respect to the output. 

In Section \ref{sec:pointconv} we illustrate more in detail KPConv and ConvPoint, with the aim of clearly presenting the differences between our composite layers and the point-convolutional layers from the literature. The major advantage of our composite layers is that we can independently customize the processing pipeline of the points coordinates (spatial information) and that of the associated features (semantic information). This design flexibility allows us to tune the hyperparameters of our composite layers to improve their performance without significantly increasing the overall number of parameters of the network. In contrast, the number of parameters of KPConv, ConvPoint, and all the other point-convolutional layers described above grows linearly with the number of centers, which is the only tunable hyperparameter of these layers.
Moreover, composite layers can aggregate spatial information and features by more general operations than linear combinations. 

Other operators similar to point-convolutions have been employed for deep learning on 3D point clouds, such as \emph{PAConv}~\cite{xu2021paconv}. In PAConv, the convolutional filters are replaced by a neural network taking as input the absolute position of the points, so the operator is not invariant to translations. Another example is \emph{MKConv}~\cite{woo2023mkconv}, where the coordinates of the points are fed to MLPs to obtain 3D tensors. These tensors are then combined with the input features and are finally processed by 3D convolutional layers to obtain the output feature vectors. Recently, also \emph{transformer}-based architectures have been applied to point clouds with promising results~\cite{fei2023dctr, yu2022point}. In our work, we do not consider these operators since we focus on defining more flexible point-convolutional layers, which are building blocks for several deep-learning solutions for point clouds.

\noindent\textbf{Graph Neural Networks.} Point clouds can also be handled by \emph{Graph Neural Networks} (GNNs)~\cite{wang2019dynamic, ma2022rethinking, wang2022novel}, which can be considered extensions of PointNet. A relevant example is \emph{Dynamic Graph CNN} (DGCNN) \cite{wang2019dynamic}, where a graph is defined by connecting some points sampled from the input point cloud to their nearest neighbors, and then processed by a GNN. Although some GNNs implement convolutional layers for graphs~\cite{wang2019dynamic, wang2022novel}, there is a substantial difference with point-convolutions since, in GNNs, the coordinates of the points are considered features of the graph vertices and thus are modified during training. In contrast, point-convolutional layers preserve the coordinates of the points and modify only the associated features.

\section{Problem Formulation} \label{sec:problem}
We define a \emph{point cloud} as a pair $( P, \phi)$, where $P \subset \mathbb{R}^d $ is a set of points in a $d$-dimensional space (in our case $d=3$), and $\phi : P \rightarrow \mathbb{R}^I$ is a function associating a feature vector $\phi(x) \in \mathbb{R}^I$ (e.g., an RGB triplet) to each point $x \in P$. Our goal is to design a new point-convolutional layer that takes as input $( P, \phi)$ and returns another point cloud $(Q, \psi)$, where $Q \subset \mathbb{R}^d $ and  $\psi : Q \rightarrow \mathbb{R}^J$. Our layer has to enable a flexible design, where we can independently set the structure and number of parameters of the components handling the spatial information and the features.

We use our layer as a building block for deep neural networks for classification, semantic segmentation, and unsupervised anomaly detection. In the latter case, the training set contains only instances from a single class, considered as \emph{normal}, and the goal during testing is detecting anomalous point clouds, i.e. those that do not belong to the normal class. This task is also called \emph{one-class classification} since only normal instances are used for training. 

\section{Background on Point Convolution}\label{sec:pointconv}
Before introducing the proposed composite layer and CompositeNets, we present some background on point convolution.

\subsection{Convolution Window and Output Point Cloud} 
Point-convolutional operators take as input a point cloud $( P, \phi)$ and return another point cloud $(Q, \psi)$. As for 2D convolutional filters, the feature vector $\psi(y)$ associated with each output point $y\in Q$ is obtained by processing a \emph{convolution window} $X_y\subset P$. While the grid structure of 2D images makes these operations straightforward, when working on point clouds there are multiple ways to define both the output point $y \in Q$ and the window $X_y \subseteq P$. 

Typically, point-convolutional layers first select the output points $\{y \in Q\}$ and then, for each $y \in Q$, define the convolution window $X_y$. A popular approach to select $y$ is randomly sampling from $P$, which implies that $Q \subseteq P$~\cite{BoulchConvpoint, AtzmonPCNNEO}. In~\cite{BoulchConvpoint}, the points are sampled with replacement from $P$, but, during sampling, points $y\in P\cap Q$ or points $y\in P$ that are close to points in $Q$ are assigned a lower probability. When applying convolution, the cardinality of the point cloud can be reduced by sampling fewer output points than there are in the input point cloud, i.e. $|Q|< |P|$, which is similar to having a stride in standard convolutions. The convolution window $X_y$ can be defined either as a sphere $X_y:= \{ x \in P : \Vert x - y\Vert < \rho \}$~\cite{HuguesKPConv}, or as a set containing a fixed number of nearest neighbors of $y$ in $P$~\cite{BoulchConvpoint}. The two methods can be considered equivalent when the input point cloud $P$ has uniform density. 

\subsection{Point-Convolutional Operators} 
After defining the output point $y$ and the convolution window $X_y$, the point-convolutional operator~\cite{HuguesKPConv, wu2019pointconv, xu2018spidercnn} is typically defined at each $y \in Q$ as:
\begin{equation}
	\psi_j(y) = (\phi * g_j)(y) = \sum_{x \in X_y} \sum_{i=1}^I \phi_i(x)g_{ij}(x -y),
	\label{eq:convolution}
\end{equation}
where $\phi_i(x)$ is the $i$-th input feature associated to $x$ and $\psi_j(y)$ is the $j$-th output feature associated to $y$. Similarly to CNNs for images, each point-convolutional layer stacks $J$ \emph{filter functions} $g_j : \mathbb{R}^d \rightarrow \mathbb{R}^{\,I}$, each having $I$ components (like the input feature function $\phi$), denoted by $g_{ij}$. 

Filters in CNNs for images are learnable weight matrices, where the size of the matrix corresponds to the size of the convolution window.
When applied to a convolution window, each weight in the filter is multiplied by the value of a specific pixel, and the weight-pixel association is determined by the grid structure of the image and the filter. In contrast, point clouds lack such grid structure, thus the filter $g_{j}$ is necessarily a function defined over $\mathbb{R}^d$. 

As stated in Section \ref{sec:related_works}, the vast majority of the point-convolutional layers in the literature operate very similarly to either KPConv~\cite{HuguesKPConv} or ConvPoint~\cite{BoulchConvpoint}, which we illustrate here more in detail. Both these layers implement each component of the filter function $g_{ij}$ as a linear combination:
\begin{equation}
	g_{ij}(x-y)=	\sum_{m=1}^M \tilde w_{ijm}H_m(x-y),\label{eq:filters}
\end{equation}
where $\tilde w_{ijm}$ are learnable weights associated to a set of spatial locations $\{c_m\}_{m=1}^M\subset\mathbb{R}^d$, called \emph{centers}, and $H_m$ are \emph{correlation functions}, each depending on the relative position of $x-y$ with respect to $c_m$~\cite{HuguesKPConv}. This formulation is inspired by convolution on images, where $x-y$ represent the pixel locations in the convolution window $X_y$, $\{c_m\}$ represent the spatial positions of the weights, and $H_m$ is the Kronecker function $H_m=\delta(x-y,c_m)$, i.e. $\delta(x-y,c_m)=1$ if $x-y=c_m$ and $0$ otherwise~\cite{BoulchConvpoint}. Due to the scattered position of the points in $X_y$, in point-convolutional layers $H_m$ must be a continuous function, and each center $c_m$ is either a fixed~\cite{HuguesKPConv} or learnable parameter~\cite{BoulchConvpoint}. In KPConv~\cite{HuguesKPConv}, each $H_m$ is an RBF, namely a function $H_m(x-y)=h(\Vert(x-y) - c_m\Vert)$ that differs from the others only by the parameter $c_m \in \mathbb{R}^d$. In ConvPoint~\cite{BoulchConvpoint}, each correlation function $H_m$ is the $m$-th output of an MLP, namely $H_m(x-y)=\text{MLP}_m \left( \left[ \left(x -y \right) -c_m\right]_{m=1}^{M} \right)$, where square brackets denote concatenation.

\begin{figure}[t]
    \centering
    \includegraphics[width=.7\columnwidth]{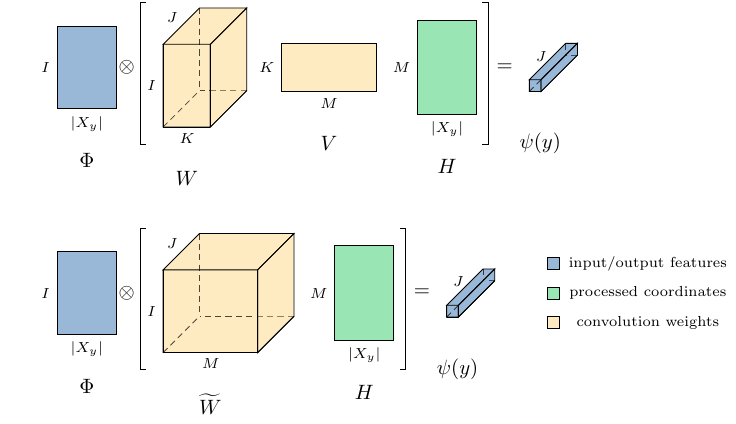}
    \begin{tikzpicture}[overlay]
	    \node at (-4.2,2.6) {\scriptsize (a) Convolutional composite layer};
	    \node at (-4.2,-0.1) {\scriptsize (b) ConvPoint / KPConv};
	\end{tikzpicture}
	\vspace{-0.3cm}
    \caption{The operations in our convolutional composite layer (a) and the well-known point-convolutional layers ConvPoint and KPConv (b) expressed in matrix form. Here $\Phi$ and $\psi(y)$ indicate, respectively, the input features of the points of the convolution window $X_y$ and the output feature vector of $y$; the matrix $H$ stacks the 3D coordinates of the points of $X_y$, processed by the correlation function $h$~\eqref{eq:H}; $\otimes$ indicates the Frobenius inner product. In practice, our composite layer decomposes the weight tensor $\widetilde{W}$ of ConvPoint and KPConv into the product $W\cdot V$, enabling more flexibility when designing the network.
    }
    \label{fig:composite-vs-kpconv}
\end{figure}

In Figure~\ref{fig:composite-vs-kpconv}(b) we illustrate the operations of the point-convolutional layers ConvPoint~\cite{BoulchConvpoint} and KPConv~\cite{HuguesKPConv} in terms of matrix multiplications. The linear combination in \eqref{eq:filters} can be expressed as a multiplication between each slice of the weight tensor $\widetilde{W}=(\tilde w_{ijm})$ and the matrix $H$ stacking the values $H_m(x-y)$ for $x\in X_y$ and $m\in\{1,\ldots,M\}$. Then, each output feature $\psi_j(y)$ can be expressed as the Frobenius inner product\footnote{namely, the sum of the entries of the element-wise product of the two matrices.} between the matrix $\Phi$ containing the input features $\phi_i(x)$ for $x\in X_y$ and $i\in\{1,\ldots,I\}$, and the $j$-th slice of $\widetilde{W}H$.

\section{Proposed Approach}\label{sec:approach}
We propose the \emph{composite layer}, a new point-convolutional operator. Our composite layer is defined by composing a \emph{spatial function} $s: \mathbb{R}^d \rightarrow \mathbb{R}^{\,K}$, which extracts the spatial information from the coordinates of the points in $X_y$, and a \emph{semantic function} $f(\phi,s)$ that combines the output of $s$ with all the input features $\{\phi(x): x\in X_y\}$ to produce the output feature vector $\psi(y)$. Therefore, as shown in Figure \ref{fig:composite-general}, our composite layer can be written as:
\begin{equation}
	\psi_j(y) = f_j(\phi, s)(y). 
	\label{eq:composite} 
\end{equation}
In what follows we show that this formulation is more general and allows superior design flexibility compared to the existing point-convolutional layers.

\subsection{Convolutional Composite Layer} \label{subsec:composite_conv}


As in ConvPoint, in our \emph{convolutional composite layer} we randomly sample each output point $y\in Q$ from the input point cloud $P$, and define the convolution windows $X_y$ by nearest neighbors. This choice guarantees that all the windows contain the same number of points, making point convolution more efficient and easier to implement.

\noindent\textbf{Spatial Function.} Inspired by KPConv~\cite{HuguesKPConv}, we implement the spatial function $s$ using a Radial Basis Function Network (RBFN) with $M$ centers, which in principle has the same approximation capability of MLPs~\cite{park1991universal} but depends on fewer hyperparameters. Similarly to the filters $g_{ij}$ in point-convolutional layers~\eqref{eq:convolution}, the $K$ components of the spatial function take as input the relative position $(x-y)$ of each point $x\in X_y$ with respect to the location of the output point $y$. We define the $k$-th component of $s$ as:
\begin{equation}
    s_k(x-y) = \sum_{m=1}^M v_{km}\,h(\Vert(x-y) - c_m\Vert),
    \label{eq:RBFN}
\end{equation}
where $v_{km}$ are learnable parameters, for $k \in\{1, ..., K\}$, and $h(r) = \exp(r^2/2\sigma^2)$, denotes a Gaussian function where $r = \Vert(x-y) - c_m\Vert$. Similarly to~\cite{AtzmonPCNNEO}, $\sigma$ is a hyperparameter common to all RBFs, and the position of the centers $\{c_m\}$ is learned during training. Let us define the matrix
\begin{equation}
    H := \begin{bmatrix}
                h(\Vert(x_1-y) - c_1\Vert) & \hdots & h(\Vert(x_{|X_y|}-y) - c_1\Vert)\\
                \vdots  & \ddots  & \vdots \\
                 h(\Vert(x_1-y) - c_M\Vert) & \hdots &  h(\Vert(x_{|X_y|}-y) - c_M\Vert)
\end{bmatrix},
    \label{eq:H}
\end{equation} which allows us to express the spatial function $s$ as a the product between a weight matrix $V=(v_{km})$ and $H$, as shown in Figure \ref{fig:composite-vs-kpconv}(a). The spatial function compresses each column of $H$ to a $K$-dimensional vector, where the hyperparameter $K$ is defined independently from the number of centers $M$ of the RBFN, and we only assume that $K<M$.

\noindent\textbf{Semantic Function.} To formulate our composite layer as a point-convolutional operator~\eqref{eq:convolution}, we define the semantic function $f$ in~\eqref{eq:composite} as: 
\begin{equation}
	\psi_j(y) = f_j(\phi, s)(y) = \sum_{x \in X_y}\sum_{i=1}^I\phi_i(x)\underbrace{\sum_{k=1}^K w_{ijk}\,s_k(x -y)}_{g_{ij}(x-y)},
	\label{eq:composite_convolution}
\end{equation}
where the indices $i,j,k$ refer to the $i$-th component of the input feature vector, the $j$-th output feature vector and the $k$-th component of the spatial function, and $\{w_{ijk}\}$ are learnable parameters. As shown in Figure \ref{fig:composite-vs-kpconv}(a), the most inner sum in \eqref{eq:composite_convolution} can be expressed as a multiplication of the projections $VH$ against each of the $J$ slices of a tensor $W=(w_{ijk})$, which contains the learnable weights of each convolution filter. Finally, each output feature $\psi_j(y)$ can be expressed as the Frobenius inner product between the matrix $\Phi$ containing the features $\phi_i(x)$ for $x\in X_y$ and $i\in\{1,\ldots,I\}$ and the $j$-th slice of the tensor $WVH$. 


\noindent\textbf{Comparison with point-convolutions.} The formulation of our composite convolution is compliant with the point-convolution definition~\eqref{eq:convolution} since the most inner sum in \eqref{eq:composite_convolution} can be interpreted as the filter function $g_{ij}$ in \eqref{eq:convolution}. 
In Figure~\ref{fig:composite-vs-kpconv} we compare the matrix operations of our convolutional composite layer with two mainstream point-convolutional layers, namely KPConv~\cite{HuguesKPConv} and ConvPoint~\cite{BoulchConvpoint}. 
When applied to a convolution window $X_y$, each of these three layers combines the input features $\Phi$ with the spatial information encoded in the matrix $H$. However, our layer uses a $K$-dimensional space (i.e. the codomain of the spatial function \eqref{eq:RBFN}) as a bottleneck between the spatial information in the matrix $H$ \eqref{eq:H} and the features\eqref{eq:composite_convolution}, while KPConv and ConvPoint directly learn a huge weight tensor $\widetilde{W}$ to combine spatial information and features. 

The most relevant advantage of our convolutional composite layer over the existing point-convolutions is a superior design flexibility: in fact, it is possible to increase the complexity (and therefore the descriptive power) of the spatial function without dramatically increasing the overall number of parameters of the layer. In particular, the complexity of the spatial function can be tuned by increasing the number of RBF centers $M$. In KPConv and ConvPoint, it is also possible to modify the number of centers $M$ to increase or decrease the complexity of the network. However, the tensor $\widetilde{W}$, which collects the vast majority of the parameters of KPConv and ConvPoint layers, grows in size linearly with $M$. In contrast, in our composite layer, only the tensor $V$, which contains substantially fewer parameters than $\widetilde{W}$, grows in size linearly with $M$ (see Figure \ref{fig:composite-vs-kpconv}). The decomposition of the weight tensor into the product $W\cdot V$ introduces an additional form of regularization compared to KPconv and ConvPoint. Assuming that $K < M$, the rank of the product $W\cdot V$ is at most $K$, since $rank(W\cdot V) = \min(rank(W),rank(V))$ and $rank(V) \leq K$. This implies that the matrix $W\cdot V$ is singular, and its rank can be adjusted by tuning a user-defined hyperparameter. Lowering the value of $K$ imposes a constraint on the possible values that each element inside $W\cdot V$ can assume, thereby reducing the layer’s expressive power in a controlled manner. Such constraints could enhance the layer's robustness, which can be particularly useful in scenarios with limited training data.

\subsection{Aggregate Composite Layer}\label{subsec:composite_aggr}
We also define an \emph{aggregate composite layer} that implements a nonlinear semantic function $f$ to aggregate the output of the spatial function $s$ \eqref{eq:RBFN} and the input features $\phi$. In particular, we adopt pooling operators defined as the component-wise mean $\mathcal{M}_s$ and standard deviation $\mathcal{S}_s$ of the output of the spatial function over the points $x\in X_y$:
\begin{eqnarray*}
    \mathcal{[M}_s(y)]_k &=& \dfrac{1}{|X_y|}\sum_{x\in X_y} s_k(x-y),\\
    \mathcal{[S}_s(y)]_k &=& \sqrt{\dfrac{\sum_{x\in X_y} (\;s_k(x-y)-[\mathcal{M}_s(y)]_k\;)^2}{|X_y|-1}}.
\end{eqnarray*}
Thus, the pooling functions $\mathcal{M}_s$ and $\mathcal{S}_s$ receive as input $|X_y|$ vectors in $\mathbb{R}^K$ (the output of the spatial function $s$), and return a single vector in $\mathbb{R}^K$ each. Analogously, we define $\mathcal{M}_\phi(y),\; \mathcal{S}_\phi(y) \in \mathbb{R}^I$ as the component-wise mean and standard deviation of the features of the points in the convolution window $X_y$. Then, we concatenate these in two vectors $\bar{s}(y) = \left[ \mathcal{M}_s(y) ; \mathcal{S}_s(y)\right]\in \mathbb{R}^{2K}$ and $\bar{\phi}(y) = \left[ \mathcal{M}_\phi(y) ; \mathcal{S}_\phi(y)\right] \in \mathbb{R}^{2I}$, and combine them as:
\begin{equation}
    f_j(\phi, s)(y) =  \bar{s}(y)^\top W_j \bar{\phi}(y) =\sum_{i=1}^{2I}\sum_{k=1}^{2K} \bar{s}_k(y) w_{ijk} \bar{\phi}_i(y),
    \label{eq:aggregate}
\end{equation}
where $W_j$ is a slice of the weight tensor $W=(w_{ijk})$. Note that the two vectors $\bar{s}(y), \bar{\phi}(y)$ are obtained by concatenating the mean and standard deviation over $X_y$, thus this semantic function is nonlinear, differently from standard point-convolution \eqref{eq:convolution}. In terms of the matrix multiplication in Figure \ref{fig:composite-vs-kpconv}(a), our aggregate composite layer combines the columns of $VH$ to extract a unique spatial descriptor for $X_y$ and does the same for $\Phi$, before computing the Frobenius product. Our aggregate layer forces the network to extract information from the mean and standard deviation of the points coordinates in $X_y$ and of the features in $\Phi$~\eqref{eq:aggregate}, thus introducing an additional form of regularization. 

\section{Experiments}\label{sec:experiments}
Our experiments are meant to show the effectiveness of composite layers in deep neural networks for point cloud classification, semantic segmentation, and, most remarkably, unsupervised anomaly detection tasks.

\subsection{Benchmarking Datasets} 
We adopt two well-known synthetic 3D classification benchmarks: \emph{ModelNet40}~\cite{WuModelnet}, which contains 12,311 objects from 40 classes, and a subset of \emph{ShapeNet}~\cite{ChangShapenet} called \emph{ShapeNetCore}, which contains 51,127 shapes from 55 rather imbalanced classes. Both datasets contain 3D meshes, from which we sample point clouds of 1024 points, setting a constant feature function $\phi\equiv1$ to construct point clouds containing only spatial information~\cite{QiPointnet, HuguesKPConv}.


We also perform our experiments on real-world data using shapes extracted from the \emph{ScanNet-v2} dataset~\cite{dai2017scannet}, which contains 1513 3D scenes (obtained from RGB-D scans) of 707 indoor environments with instance-level annotations of 608 classes. Specifically, we extract the meshes of objects belonging to the 17 most populated classes (excluding planar objects such as windows, walls, and doors) as in the classification experiments in~\cite{dai2017scannet}, and sample 1024 points from each object, as for the synthetic datasets. 
The resulting dataset is severely imbalanced. Moreover, the objects are often partially occluded since they were acquired from real-world scenes. 
Finally, we use the real-world segmentation benchmark \emph{S3DIS}~\cite{armeni20163d}, which contains 5 large-scale 3D scenes (obtained from RGB-D scans) from 3 buildings with semantic annotations for 12 classes of structural elements and pieces of furniture.

For classification, we train each network using all the classes of the official training split of ModelNet40, ShapeNetCore, and ScanNet. For segmentation, we use the first 4 scenes of S3DIS for training and the fifth one for testing, as in \cite{HuguesKPConv}. For anomaly detection, we consider exclusively point clouds from the 7 most represented classes from ShapeNet (at least 2000 instances, as in \cite{masuda2021toward}) and the 14 largest classes from ScanNet (at least 400 instances), using the same training/test splits as for classification. In these experiment, the networks are trained in a self-supervised fashion, i.e. using only instances from a single class that is considered normal. During testing, point clouds from any other class are considered anomalous. 





\begin{table}[t]
\centering
\renewcommand{\arraystretch}{0.75}
\setlength{\tabcolsep}{12pt}
\caption{The hyperparameters of our CompositeNets: for each layer, $J$ is the number of output features, expressed as a function of a dataset-specific parameter $J_0$, $|X_y|$ is the cardinality of each convolution window $X_y$, and $|Q|$ is the cardinality of the output point cloud $Q$.}\label{tab:MCarch}
\resizebox{\textwidth}{!}{  
\begin{tabular}{c|ccc|c|ccc}
\toprule
\multicolumn{4}{c|}{CompositeNet architecture} & \multicolumn{4}{c}{U-Net CompositeNet architecture (encoder)}\\
\midrule
\textbf{Layer Type}       & $J$     & $|X_y|$ & $ |Q| $ & \textbf{Layer Type}       & $J$     & $|X_y|$ & $ |Q| $ \\
\midrule
Composite + BN + ReLU & $J_0$   & 32      & 1024 & Composite + BN + ReLU & $J_0$   & 16& 8192   \\
\midrule
Composite + BN + ReLU & $2\cdot J_0$ & 32      & 256 & Composite + BN + ReLU& $J_0$   & 16&2048    \\
\midrule
Composite + BN + ReLU & $4\cdot J_0$ & 16      & 64 & Composite + BN + ReLU & $J_0$   & 16& 1024     \\
\midrule
Composite + BN + ReLU & $4\cdot J_0$ & 16      & 16 & Composite + BN + ReLU & $J_0$   & 16      & 256     \\
\midrule
Composite + BN + ReLU & $8\cdot J_0$ & 16      & 1 & Composite + BN + ReLU & $2\cdot J_0$& 8& 64      \\
\midrule
Dense            & --      & --      & -- & Composite + BN + ReLU & $2\cdot J_0$& 8&16   \\
\midrule
\multicolumn{4}{c|}{ } & Composite + BN + ReLU & $2\cdot J_0$& 4& 8\\
\bottomrule
\end{tabular}
}
\end{table}


\subsection{Architectural Details}
Our CompositeNet for classification and anomaly detection, which we indicate by $\mathcal{N}$, is a simple neural network consisting of 5 composite layers. Our architecture is illustrated in Table~\ref{tab:MCarch}, which reports the number of output features $J$, the window cardinality $|X_y|$, and the number of output points $|Q|$ for each layer. We chose these parameters so that our architecture is equivalent to the deep neural network based on ConvPoint layers in~\cite{BoulchConvpoint}. The number of output features $J$ of each layer is proportional to the number of output features $J_0$ of the first layer, which is a hyperparameter. We apply nonlinearities and batch normalization only to the features and not to the spatial coordinates of the points: this is similar to what happens in CNNs for images, where the spatial coordinates of pixels are not modified. The cardinality of the point cloud is reduced throughout the network by sampling, in each layer, fewer output points than input points. The last composite layer returns a single point with a feature vector, which is fed to a dense layer, discarding its coordinates. Using the architecture illustrated in Table~\ref{tab:MCarch}, we implement a convolutional CompositeNet based on our convolutional composite layer (Section \ref{subsec:composite_conv}) and an aggregate one based on our aggregate composite layer (Section \ref{subsec:composite_aggr}). 

For semantic segmentation, we use our CompositeNets to implement the \emph{fusion} architecture proposed in~\cite{BoulchConvpoint}. Our model consists of two U-Nets based on CompositeNet, one receiving RGB input and the other only the spatial information. The structure of the U-Net encoder is shown in Table \ref{tab:MCarch}. Then, we concatenate the features obtained before the fully-connected layer from both U-Nets and feed them to a \emph{fusion module} composed of two composite layers and a fully-connected one. The final output is obtained by summing the outputs of the two initial U-Nets and the fusion module. In our preliminary experiments, we also used U-Nets based on the same sequential architecture we used for classification, achieving inferior results, in line with those obtained by ConvPoint with the same architecture~\cite{BoulchConvpoint}.



\begin{table}[]
\centering
\renewcommand{\arraystretch}{0.75}
\setlength{\tabcolsep}{12pt}
\caption{Hyperparameters of our convolutional and aggregate CompositeNets, ConvPoint and KPConv-vanilla maximizing the validation accuracy/mIoU on each considered dataset: $J_0$ regulates the number of output features of each layer, $M$ is the number of RBF centers for our CompositeNets and KPConv-vanilla and the size of the MLP output in ConvPoint, while $K$ is the size of the spatial function output in our CompositeNets.}\label{tab:MCHP}

                              
\resizebox{\textwidth}{!}{
\begin{tabular}{c|ccc|ccc|ccc|ccc}
\toprule
& \multicolumn{3}{c|}{ModelNet40~\cite{WuModelnet}} & \multicolumn{3}{c|}{ShapeNetCore~\cite{ChangShapenet}} & \multicolumn{3}{c|}{ScanNet~\cite{dai2017scannet}} & \multicolumn{3}{c}{S3DIS~\cite{armeni20163d}}\\
\midrule
\textbf{Layer}                      & $J_0$       & $M$        & $K$ & $J_0$       & $M$        & $K$ & $J_0$       & $M$        & $K$ & $J_0$       & $M$        & $K$  \\
\midrule
\textbf{Ours (Conv.)} & 64          & 64         & 16 & 32          & 64         & 16  & 64          & 256        & 32  & 64 &  128 & 16     \\
\textbf{Ours (Aggr.)} & 64          & 64         & 16 & 32          & 64         & 16 & 64          & 256        & 32    & 64 &  128 & 16  \\
\textbf{ConvPoint}    & 64          & 16         & -- & 32          & 16         & -- & 64          & 32         & --     & 64 &  16 & --  \\
\textbf{KPConv-vanilla}    & 64          & 16         & -- & 64          & 32         & -- & 64          & 64         & -- \\
\bottomrule
\end{tabular}
}
\end{table}

\subsection{Multiclass Classification} 
We train our CompositeNets for 200 epochs on an NVIDIA RTX A6000 GPU, using the Adam optimizer \cite{kingma2014adam} and the cross-entropy loss. In Table \ref{tab:MCHP} we report the hyperparameter values maximizing the validation accuracy.

\noindent\textbf{Considered methods.}
We compare our CompositeNets against PointNet~\cite{QiPointnet}, KPConv~\cite{HuguesKPConv}, and ConvPoint~\cite{BoulchConvpoint}, reporting their accuracy on ModelNet40 from the corresponding papers. Since the performance on ShapeNetCore and ScanNet was not reported, we train and test the competing networks using the software implementations made publicly available. We remark that both our CompositeNets are equivalent to ConvPoint in terms of architecture, which makes the comparison between our CompositeNets and ConvPoint particularly significant. In contrast, PointNet uses a custom non-convolutional architecture and KPConv uses a deeper residual network, which are both significantly different from ours. We also train \emph{KPConv-vanilla}, a network based on KPConv with the same architecture as our CompositeNets. In Table~\ref{tab:MCHP} we report the configuration of ConvPoint proposed in \cite{BoulchConvpoint} for ModelNet40, and the configurations we selected for ConvPoint and KPConv-vanilla to maximize the validation accuracy also on the other datasets. KPConv layers have an additional hyperparameter compared to ConvPoint, namely the radius $\rho$ defining the convolution windows~\cite{HuguesKPConv}. In KPConv-vanilla we have tuned $\rho$ in a $\pm$20\% range compared to the value in~\cite{HuguesKPConv} without obtaining substantial changes in the accuracy.

\noindent\textbf{Figures of merit.} We consider two performance metrics: the overall accuracy (OA), defined as the proportion of correctly classified point clouds, and the average accuracy (AA), namely the average per-class accuracy, which better takes into account the accuracy on under-represented classes. We train each method 5 times and compute the average OA and AA, except for the results on ModelNet40, which we report from \cite{QiPointnet, HuguesKPConv, BoulchConvpoint}.

\begin{table*}[t!]
\caption{Classification performance (OA and AA). The results of competing methods on ModelNet40 are from the literature (KPConv~\cite{HuguesKPConv} does not report the AA), while those on ShapeNetCore and ScanNet are obtained using the available code.}\label{tab:MC} 
\centering
\renewcommand{\arraystretch}{0.75}
\setlength{\tabcolsep}{12pt}
\resizebox{\textwidth}{!}{  
\begin{tabular}{c  | c c    |cc |c c    | c c |     c c |    c c }
\toprule
&  \multicolumn{2}{c|}{\textbf{PointNet}~\cite{QiPointnet} } & \multicolumn{2}{c|}{\textbf{KPConv}~\cite{HuguesKPConv}} & \multicolumn{2}{c|}{\textbf{KPConv-vanilla}} & \multicolumn{2}{c|}{\textbf{ConvPoint}~\cite{BoulchConvpoint}} & \multicolumn{2}{c|}{\textbf{Ours (Conv.)}} & \multicolumn{2}{c}{\textbf{Ours (Aggr.)}} \\
\midrule
 \textbf{Dataset}   & OA   & AA  & OA & AA & OA & AA & OA & AA & OA & AA & OA & AA\\
\midrule
ModelNet40~\cite{WuModelnet} & 0.892    & 0.862   & \textbf{0.929}   & --  & 0.879 & 0.848 & 0.918   &  \textbf{0.885}   & 0.913  & 0.871   & 0.911   & 0.879 \\
ShapeNetCore~\cite{ChangShapenet} & 0.830    & 0.686   & 0.829   & 0.671  & 0.752  & 0.632  & 0.837   & 0.669   & 0.828   & 0.653   & \textbf{0.839}   & \textbf{0.698} \\
ScanNet~\cite{dai2017scannet} & 0.786  & 0.751  & \textbf{0.878}  & 0.835 & 0.766 & 0.699  & 0.847  &  0.811  & 0.861   & 0.824   & 0.871 & \textbf{0.839} \\
\bottomrule
\end{tabular}
}
\end{table*}

\noindent\textbf{Experimental results.}
The results in Table~\ref{tab:MC} indicate that our CompositeNets achieve comparable performance to KPConv~\cite{HuguesKPConv} and ConvPoint~\cite{BoulchConvpoint} on all the considered datasets. Therefore, the point-convolutional layers in these models can be successfully replaced by our composite layers. Moreover, the poor performance of KPConv-vanilla confirms that, unlike ours, KPConv layers require a deeper architecture with residual blocks to achieve a good accuracy.

On ModelNet40~\cite{WuModelnet}, our CompositeNets approach ConvPoint~\cite{BoulchConvpoint} both in terms of OA and AA, and substantially outperform PointNet~\cite{QiPointnet} in terms of OA. KPConv has the best OA (its AA is not reported in~\cite{HuguesKPConv}). On ShapeNetCore~\cite{ChangShapenet}, our aggregate CompositeNet yields the best OA, slightly outperforming ConvPoint, while our convolutional CompositeNet performs similarly to PointNet and KPConv. Our aggregate CompositeNet has also the best AA, which is particularly important on imbalanced datasets such as ShapeNetCore. On ScanNet~\cite{dai2017scannet} our aggregate CompositeNet performs very similarly to KPConv, which has a slightly better OA. Also in this case, our aggregate CompositeNet yields the best AA, suggesting that our aggregate CompositeNet is more robust to class imbalance than other methods. Remarkably, on challenging datasets such as ShapeNetCore and ScanNet our aggregate CompositeNet outperforms ConvPoint, which shares the same sequential architecture, and achieves similar accuracy to KPConv, which has more layers and residual blocks. 
In most cases, our aggregate CompositeNet performs better than ConvPoint, which is in line with our convolutional CompositeNet.


\subsection{Semantic Segmentation}
We train our fusion CompositeNets for 50 epochs on an NVIDIA RTX A6000 GPU, using the Adam optimizer \cite{kingma2014adam} and the cross-entropy loss. As recommended in \cite{BoulchConvpoint}, we perform two separate steps: first, we train the U-Nets independently; then, we load the weights of the two U-Nets on the fusion module and train it. During the second step, the weights of the two U-Nets continue to be updated. In Table~\ref{tab:MCHP} we report the hyperparameters maximizing the mIoU on the validation set.

\begin{table*}[t!]
\caption{Semantic segmentation performance (IoU) over each class of the S3DIS dataset~\cite{armeni20163d}. Our aggregate CompositeNet is the best-performing method in terms of average rank, while KPConv~\cite{HuguesKPConv} has the best mIoU. KPConv~\cite{HuguesKPConv} does not report the OA.}\label{tab:s3dis} 
\centering
\renewcommand{\arraystretch}{0.75}
\setlength{\tabcolsep}{12pt}
\resizebox{0.75\textwidth}{!}{  
\begin{tabular}{c|c|c|c|c}
\toprule
\textbf{Class} & \textbf{KPConv}~\cite{HuguesKPConv} & \textbf{ConvPoint}~\cite{BoulchConvpoint} & \textbf{Ours (Conv.)} & \textbf{Ours (Aggr.)} \\
\midrule
Beam & 0.000 & 0.000 & 0.000 & 0.000\\
Board & 0.637 & 0.478 & 0.501& \textbf{0.664}\\
Bookcase & 0.746 & 0.679 & 0.716& \textbf{0.773}\\
Ceiling & 0.926 & \textbf{0.978}& 0.954& 0.948\\
Chair & 0.901 & 0.881 & 0.840& \textbf{0.903}\\
Clutter & \textbf{0.581} & 0.539 & 0.549& 0.556 \\
Column & 0.165 & 0.169 & \textbf{0.276}& 0.229\\
Door & \textbf{0.695} & 0.173 & 0.557& 0.175\\
Floor & 0.973 & \textbf{0.985} & 0.954& 0.976\\
Sofa & \textbf{0.664} & 0.272 & 0.485& 0.480\\
Table & 0.802 & \textbf{0.837} & 0.770& 0.817\\
Wall & 0.814 & \textbf{0.822} & 0.801& 0.811\\
Window & \textbf{0.545} & 0.378 & 0.464& 0.496\\
\midrule
\textbf{Avg. Rank} & 2.167 & 2.833 & 2.917 & \textbf{2.083} \\
\midrule
\textbf{Wilcoxon-p} & 5.45 $\cdot$ 10\textsuperscript{--1} & 4.61 $\cdot$ 10\textsuperscript{--2} & 1.17 $\cdot$ 10\textsuperscript{--1} & -- \\
\midrule
\textbf{mIoU} & \textbf{0.654} & 0.553& 0.603& 0.602\\
\textbf{OA} & -- & 0.851 & \textbf{0.869}& 0.861\\

\bottomrule
\end{tabular}
}
\end{table*}


\noindent\textbf{Considered methods.}
We compare our CompositeNets against KPConv~\cite{HuguesKPConv} and ConvPoint~\cite{BoulchConvpoint} on the S3DIS dataset~\cite{armeni20163d}. For KPConv, we report the results presented in~\cite{HuguesKPConv}, while we retrain ConvPoint with the same training/testing split as our CompositeNets using the available code.

\noindent\textbf{Figures of merit.} We assess segmentation performance by the Intersection over Union (IoU) over each class. Following~\cite{demvsar2006statistical}, we report the average rank of each method over the classes and the p-values of the one-sided Wilcoxon Signed-Rank test assessing whether the performance difference between the best-ranking method and the others is statistically significant. We also report the mean IoU over all classes (mIoU) and the OA.

\noindent\textbf{Experimental results.} The results in Table~\ref{tab:s3dis} show that our aggregate CompositeNet is the best-performing method in terms of average rank, and the difference with ConvPoint is statistically significant (p-value $\leq0.05$) according to the Wilcoxon test. Our convolutional CompositeNet has a slightly worse average rank than ConvPoint, but both our CompositeNets outperform ConvPoint in terms of mIoU and OA. KPConv has a slightly worse average rank than our aggregate CompositeNet, but has a better mIoU due to the relatively worse performance of our CompositeNets on specific classes such as \emph{Door} and \emph{Sofa}. None of the considered methods can identify \emph{Beam} since it is an under-represented class. These results confirm the trends observed in classification.

\begin{table*}[t!]
\centering
\caption{Anomaly detection performance (AUC) of our self-supervised CompositeNets, two similar networks based on ConvPoint~\cite{BoulchConvpoint} and KPConv~\cite{HuguesKPConv}, and IFOR~\cite{liu2008isolation} applied to the GOOD descriptor \cite{kasaei2016good}. We also report the results obtained by VAE~\cite{masuda2021toward} on ShapeNet. Our aggregate CompositeNet is the best method in most classes and in terms of average rank.
}\label{tab:shallow}
\renewcommand{\arraystretch}{0.75}
\setlength{\tabcolsep}{12pt}
\resizebox{\textwidth}{!}{  
\begin{tabular}{cc|c|c|cccc} 
\toprule
& & Baseline   & SOTA & \multicolumn{4}{c}{Self-supervised deep neural networks}\\
\midrule
& \textbf{Class}                        & \textbf{IFOR } & \textbf{VAE}~\cite{masuda2021toward}         & \textbf{KPConv} & \textbf{ConvPoint} & \textbf{Ours (Conv.)}  & \textbf{Ours (Aggr.)}   \\ 
\midrule
\multirow{7}{*}{\rotatebox{90}{ShapeNet~\cite{ChangShapenet}}} 
& Airplane & 0.912 & 0.747 & \textbf{0.974} & 0.956 & 0.969 & 0.970\\
& Car & 0.712 & 0.757 & \textbf{0.988} & 0.953 & 0.953 & 0.972\\
& Chair & 0.571 & 0.931 & 0.921 & 0.910 & 0.904 & \textbf{0.941}\\
& Lamp & \textbf{0.962} & 0.907 & 0.372 & 0.373 & 0.391 & 0.424\\
& Table & 0.883 & 0.839 & \textbf{0.943} & 0.777 & 0.821 & 0.854\\
& Rifle & 0.475 & 0.382 & 0.967 & 0.976 & \textbf{0.978} & 0.977\\
& Sofa &  \textbf{0.986} & 0.777 & 0.923 & 0.898 & 0.936 & 0.944 \\
\midrule
& \textbf{Avg. Rank}    & 3.714 & 4.429 & 2.857 & 4.357 & 3.500 & \textbf{2.143}                    \\ 
\midrule
& \textbf{Wilcoxon-p}   & 2.89 $\cdot$ 10\textsuperscript{--1} & 1.09 $\cdot$ 10\textsuperscript{--1} & 3.44 $\cdot$ 10\textsuperscript{--1} & 7.81 $\cdot$ 10\textsuperscript{--3} & 2.34 $\cdot$ 10\textsuperscript{--2} & \textbf{--}                 \\

\midrule
\multirow{14}{*}{\rotatebox{90}{ScanNet~\cite{dai2017scannet}}} 
& Backpack              & 0.677 & --   & \textbf{0.851}                  & 0.778         & 0.783                & 0.779\\
& Book                  & \textbf{0.830} & --  & 0.734          & 0.731         & 0.782                         & 0.782\\
& Bookshelf             & 0.745 & --   & 0.796                  & 0.792          & \textbf{0.824}               & 0.798\\
& Box                   & 0.429 & --   & \textbf{0.705}                    & 0.638         & 0.630                         & 0.659\\
& Cabinet               & 0.581 & --   & 0.813                   & 0.818          & 0.810                & \textbf{0.827}\\
& Chair                 & 0.449 & --   & 0.747                      & \textbf{0.842}          & 0.837               & 0.838\\
& Desk                  & 0.482 & --   & 0.869                      & 0.857          & 0.870               & \textbf{0.888}\\
& Lamp                  & 0.540 & --   & 0.639                       & 0.707          & 0.703               & \textbf{0.725} \\
& Pillow                & 0.647 & --   & 0.693                    & 0.707          & 0.754               & \textbf{0.771} \\
& Sink                  & 0.415 & --   & 0.896                       & 0.892        & 0.921                & \textbf{0.940}  \\
& Sofa                  & 0.785 & --   & \textbf{0.876}              & 0.859          & 0.852               & 0.852\\
& Table                 & 0.318 & --   & \textbf{0.920}                      & 0.912          & 0.902               & 0.902\\
& Towel                 & \textbf{0.801} & --   & 0.657             & 0.690                   & 0.640               & 0.655\\
& Trash Can             & 0.426 & --   & 0.881                 & 0.918          & 0.921               & \textbf{0.927}\\
\midrule
& \textbf{Avg. Rank}    & 4.727 & -- & 2.000 & 3.227 & 3.273 & \textbf{1.773}                    \\ 
\midrule
& \textbf{Wilcoxon-p}   & 1.46 $\cdot$ 10\textsuperscript{--3} & -- & 6.50 $\cdot$ 10\textsuperscript{--1} & 2.53 $\cdot$ 10\textsuperscript{--3} & 9.28 $\cdot$ 10\textsuperscript{--3} & \textbf{--}                 \\

\bottomrule
\end{tabular}
}
\end{table*}

\subsection{Unsupervised Anomaly Detection} 
We train our CompositeNets for anomaly detection using a self-supervised approach~\cite{golan2018deep}, which we briefly illustrate here. Starting from a training set containing only point clouds from a given normal class, we form a new self-annotated dataset by applying $N$ different geometric transformations $\{\mathcal{T}_0,\ldots,\mathcal{T}_{N-1}\}$ to the point clouds of the training set, where $\mathcal{T}_0$ is the identity. Then, we train a CompositeNet $\mathcal{N}$ to classify the transformations applied to each input point cloud. The rationale is that, by learning the auxiliary task of distinguishing transformations, the network also learns features that characterize the normal class~\cite{golan2018deep}. During testing, we apply the transformations to each point cloud $P$ and classify the resulting $\mathcal{T}_n(P)$. Then, we use as normality score the average posterior returned by $\mathcal{N}$ for the correct transformations applied to $P$, i.e.
\begin{equation}\label{eq:normalityscore}
    S(P) = \dfrac{1}{N}\sum_{n=0}^{N-1} \big[ \mathcal{N}\big(\mathcal{T}_n(P)\big)\big]_n.
\end{equation}
This method was designed for 2D images, so the geometric transformations included rotations, translations, and flips~\cite{golan2018deep}, but translations cannot be directly adopted on 3D point clouds since neural networks for point clouds are translation-invariant. Moreover, most real-world objects (e.g. chairs, tables, etc.) have a common vertical orientation and thus appear arbitrarily rotated around the vertical axis in the training set. For this reason, we employ a set of $N=8$ rotations of an angle $\alpha\in\{0^{\circ},45^{\circ},90^{\circ},135^{\circ},210^{\circ},240^{\circ},300^{\circ},330^{\circ}\}$ around a fixed horizontal axis.
We train our self-supervised CompositeNets for 200 epochs on an NVIDIA RTX A6000 GPU using the Adam optimizer~\cite{kingma2014adam} and the same hyperparameters selected for our classification experiments (Table \ref{tab:MCHP}).

\noindent\textbf{Considered methods.} Only a few works perform unsupervised learning on point clouds in applications such as detecting defective products in manufacturing~\cite{lyu2021online}, recognizing vehicles in autonomous driving~\cite{qian20223d} or identifying geometric structures~\cite{ao2017one}. These methods leverage hand-crafted features, also referred to as \emph{point cloud descriptors} \cite{hana2018comprehensive}. For this reason, we consider a baseline leveraging the \emph{Global Ortographic Object Descriptor} (GOOD) \cite{kasaei2016good}. Compared to most of the existing descriptors~\cite{hana2018comprehensive}, the advantage of GOOD is that it does not require additional point cloud features such as the normal vectors to the surface at each point. To perform anomaly detection, we first use GOOD to extract a feature vector from each normal point cloud in the training set. Then, we feed these feature vectors to an \emph{Isolation Forest} (IFOR)~\cite{liu2008isolation}, a well-known outlier-detector.

To the best of our knowledge, the only work using deep learning for anomaly detection in point clouds is \cite{masuda2021toward}, in particular, a variational autoencoder (VAE). Since the implementation is not publicly available, we report from \cite{masuda2021toward} the results obtained on the 7 most numerous classes of the ShapeNet dataset. We also compare our self-supervised CompositeNets with KPConv~\cite{HuguesKPConv} and ConvPoint~\cite{BoulchConvpoint}, trained in the same fashion using the same architectures as in classification. 
Another work addressing anomaly detection in point clouds leverages a GNN that was pre-trained for segmentation on both normal and anomalous point clouds~\cite{gencer2022one}, so it is not unsupervised and thus we do not consider it.


\noindent\textbf{Figures of merit.} We assess the anomaly detection performance by the Area Under the ROC Curve (AUC), which measures how well each network can distinguish normal instances from anomalous ones. Following~\cite{demvsar2006statistical}, we report the average rank of each method over the classes and the p-values of the one-sided Wilcoxon Signed-Rank test assessing whether the performance difference between the best-ranking method and the others is statistically significant. 

\noindent\textbf{Experimental results.} In Table \ref{tab:shallow} we report the results of our experiments. 
On ShapeNet, self-supervised deep neural networks outperform both the shallow baseline IFOR and the VAE~\cite{masuda2021toward} in most of the classes and in terms of average rank. However, the Wilcoxon test~\cite{demvsar2006statistical} comparing our aggregate CompositeNet (which is the best method in terms of average rank) and these baselines yields p-value $>0.05$, probably due to the fact that the number of considered classes is small. Unfortunately, we could not compare our CompositeNets with~\cite{masuda2021toward} on more classes since its implementation is not publicly available.

On ScanNet, all methods perform worse than on the synthetic data from ShapeNet. We expected this result since the real-world objects in ScanNet are often partially occluded. Also in this case, we observe that self-supervised deep neural networks outperform IFOR in most of the classes. Our aggregate CompositeNet is the best-performing method in most classes and in terms of average rank, and the Wilcoxon test~\cite{demvsar2006statistical} confirms that the difference with all the other methods, except for KPConv, is statistically significant (p-value $\leq0.05$). These results show that, in both the considered datasets, our aggregate CompositeNet substantially outperforms ConvPoint, which has the same sequential architecture, and performs similarly to KPConv, which instead uses residual blocks.  

Finally, we observe that applying IFOR to feature vectors extracted by GOOD yields relatively good performance on a small number of classes, outperforming deep solutions on the \emph{Lamp} and \emph{Sofa} classes from ShapeNet, and on the \emph{Book} and \emph{Towel} classes from ScanNet. However, IFOR yields poor performance on most of the other classes, often achieving AUC $<0.5$. This suggests that point cloud descriptors cannot characterize well all classes of point clouds, thus deep learning can address anomaly detection in a more general way.

\subsection{Ablation Studies}\label{subsec:flexibility}
\begin{figure*}[t]
    \centering
    \begin{tikzpicture}
\begin{groupplot}[ 
    group style={group size=2 by 1, vertical sep=1.2cm,
    horizontal sep=1.2cm},
    width=0.40\textwidth,
    height=0.276\textwidth,
    title style = {font=\small,
    },
    scale only axis,
    ymajorgrids,
    xlabel style = {anchor = near ticklabel, 
        	at = {(0.5,-0.1)},font=\footnotesize},
    ylabel style = {at={(-0.12,0.5)},
            font=\footnotesize},
    xticklabel style = {font = \footnotesize},
    yticklabel style = {font = \footnotesize},
    y tick label style={/pgf/number format/.cd,1000 sep={},scale only axis,scaled ticks=false,
    fixed zerofill,
    precision=2}
]
%
\nextgroupplot[
        xticklabels = {$2$,$4$,$8$,$16$,$32$,$64$},
        xmode = log,
        log ticks with fixed point,
        xtick={2e6,4e6,8e6,16e6,32e6,64e6},
        xlabel={\#parameters ($\cdot10^6$)},
        ylabel={overall accuracy},
        title={Comparison with ConvPoint and KPConv-vanilla},
        ]
name=ax1,
scale only axis,
xmode = log,
title style = {
    at = {(0.5, 0.97)},
},
xlabel style = {
	align=center,
	anchor = near ticklabel, 
	at = {(0.5,-0.1)},
},
ylabel style = {
    anchor = center, 
    at = {(-0.12,0.5)},
    font=\small,
},

ymax = 0.875,
ytick = {0.65,0.7,...,0.9},
ymajorgrids=true,
legend style={at={(0.99, 0.01)},
	anchor=south east, 
	font=\scriptsize,
	legend cell align=left,
	legend columns=2},
y tick label style={
    font=\small,
    /pgf/number format,
    fixed,
    fixed zerofill,
    precision=2}
]
\coordinate (insetPosition) at (rel axis cs:1,-0.005);
\coordinate (c1) at (rel axis cs:0.51,0.535);
\coordinate (c2) at (rel axis cs:0.645,0.93);
\addplot[style_convpoint]
table [
    x=convpointP,
    y=convpoint,
]{data/params.txt};
%
\addplot[style_kpconv, dashed, mark options={solid}]
table [
    x=kpconvP,
    y=kpconv,
]{data/params.txt};

\addplot[style_linear]
table [
    x=linearP,
    y=linear,
]{data/params.txt};
%
\addplot[style_aggr]
table [
    x=aggrP,
    y=aggr,
]{data/params.txt};
\node [below] at (axis cs:2000000,0.785) {\scriptsize$0.56\text{ms}$};
\node [] at (axis cs:64000000,0.858) {\scriptsize$2.85\text{ms}$};
\node [right] at (axis cs:4000000,0.775) {\scriptsize$0.67\text{ms}$};
\node [right] at (axis cs:4000000,0.847) {\scriptsize$1.74\text{ms}$};
\node [right] at (axis cs:15500000,0.858) {\scriptsize$1.78\text{ms}$};
\node [right] at (axis cs:15500000,0.824) {\scriptsize$0.70\text{ms}$};
\node [above] at (axis cs:1500000,0.70) {\scriptsize$0.90\text{ms}$};
\node [above] at (axis cs:45000000,0.75) {\scriptsize$3.32\text{ms}$};


\nextgroupplot[
        xmax=17,
        xmin= 3,
        ymax = 0.94, ymin = 0.87,
        x dir=reverse,
        xtick={4,8,12,16},
        ytick={0.88,0.89,...,0.93},
        xticklabels = {$4$,$8$,$12$,$16$},
        xlabel={$K$ ($M$ for KPConv)},
        title={Comparison with KPConv},
        legend style={at={($(0,0)+(0cm,0cm)$)},legend columns=-1,fill=none,draw=black,anchor=center,align=center, font=\scriptsize, legend to name = grouplegend,}
        ]
\addplot[style_convpoint]
coordinates{(0,0.89) (0,0.9)};
\addlegendentry{ConvPoint}
\addplot[style_kpconv, dashed, mark options={solid}]
coordinates{(12,0.7)(16,0.7)};
\addlegendentry{KPConv-vanilla}
\addplot[style_kpconv]
table [
    x=KPConvP,
    y=KPConv,
]{data/tuningK.txt};
\addlegendentry{KPConv}
\addplot[style_linear]
table [
    x=linearP,
    y=linear,
]{data/tuningK.txt};
\addlegendentry{Ours (Conv.)}
\addplot[style_aggr]
table [
    x=aggrP,
    y=aggr,
]{data/tuningK.txt};
\addlegendentry{Ours (Aggr.)}

\node [below] at (axis cs:4.4,0.878) {\scriptsize$3.9\cdot10^6$};
\node [above] at (axis cs:15.6,0.93) {\scriptsize$6.3\cdot10^6$};
\node [above] at (axis cs:4.4,0.906) {\scriptsize$3.8\cdot10^6$};
\node [above] at (axis cs:15.6,0.915) {\scriptsize$1.5\cdot10^7$};
\node [below] at (axis cs:4.4,0.897) {\scriptsize$9.8\cdot10^5 $};
\node [below] at (axis cs:15.6,0.91) {\scriptsize$3.9\cdot10^6$};

\end{groupplot}
\node[align=center] at ($(group c1r1.south west)!.5!(group c1r1.south east) + (0cm,-1.05cm)$) {\scriptsize(a)};
\node[align=center] at ($(group c2r1.south west)!.5!(group c2r1.south east) + (0cm,-1.05cm)$) {\scriptsize(b)};
\node at ($(group c1r1)!.5!(group c2r1) + (0cm,-3.2cm)$) {\ref{grouplegend}}; 
\end{tikzpicture}%
    \caption{(a) Overall accuracy on ScanNet against the number of parameters, varying the number of centers $M$. We report the average processing time during training for the least and most complex versions. The parameters of ConvPoint and KPConv-vanilla increase linearly with $M$, while in our CompositeNets $M$ can be increased without significantly changing the number of parameters. (b) Overall accuracy on ModelNet40 when reducing the number of parameters by tuning $K$ ($M$ for KPConv). We also report the number of parameters of the most and least complex versions. These results show that we can substantially reduce the parameters of our CompositeNets without compromising the accuracy, while this is not the case in KPConv.}
    \label{fig:param-tuning}
\end{figure*}

We perform a few ablation studies, to investigate the superior design flexibility of our composite layers compared to KPConv~\cite{HuguesKPConv} and ConvPoint~\cite{BoulchConvpoint}. First, we show that in CompositeNets we can easily customize the spatial and semantic components of the composite layer to \emph{i)} improve the classification performance without significantly increasing the overall number of parameters, and \emph{ii)} reduce the number of parameters without compromising performance. In KPConv and ConvPoint, similar adjustments imply either a significant increase in the number of parameters or a performance drop. 
Then, we investigate the advantages of combining convolutional and aggregate layers in the same architecture.

\noindent\textbf{Increasing the complexity of the spatial function.} In the first experiment, we train different configurations of our CompositeNets on the ScanNet dataset~\cite{dai2017scannet}, modifying the number of centers $M\in\{8,16,\ldots,256\}$ to modify the number of parameters of the spatial function. To enable a fair comparison, we designed our CompositeNets and a KPConv-vanilla model to match the architecture of ConvPoint. In ConvPoint and KPConv-vanilla, we tune $M\in\{8,16,\ldots,256\}$ as in our CompositeNets. Note that \emph{KPConv-vanilla} has a less deep architecture without residual blocks compared to that in \cite{HuguesKPConv}. 

As remarked in Section \ref{subsec:composite_conv}, in ConvPoint and KPConv layers the number of parameters increases linearly with $M$, as this hyperparameter determines the size of the weight tensor $\widetilde{W}$ (see Figure \ref{fig:composite-vs-kpconv}). In contrast, in our composite layers we can modify the complexity of the spatial function only by adjusting $M$ and keeping the output dimension $K=16$ fixed. Changing $M$ only modifies the size of the weight matrix $V$, while keeping unaltered the sizes of the weight tensor $W$, which contains most of the learnable parameters. 

Figure \ref{fig:param-tuning}(a) shows the overall accuracy (OA) over the ScanNet dataset against the number of parameters of the networks. In ConvPoint and KPConv-vanilla the number of parameters grows linearly with $M$, while in our CompositeNets it increases less than 1\% when passing from $M=8$ to $M=256$. In our CompositeNets, the accuracy increases with $M$, while ConvPoint and KPConv-vanilla achieve their best performance at $M=64$ and yield worse accuracy at $M=128$ and $256$, suggesting that they might be more prone to overfitting due to the increased number of parameters. 

In Figure \ref{fig:param-tuning}(a) we also report the average processing time of a point cloud while training the least and most complex versions of each network. We observe that, in all cases, the processing time increases with $M$ due to the increasing complexity of the spatial component. However, in ConvPoint and KPConv-vanilla the processing time increases more than in our CompositeNets due to the linear growth of the number of parameters with respect to $M$.

We observe a similar trend on the synthetic ModelNet40~\cite{WuModelnet} and ShapeNetCore~\cite{ChangShapenet} datasets, although the performance difference when varying $M\in\{64,128,256\}$ is not as evident as in ScanNet~\cite{dai2017scannet}. Therefore, on ModelNet40 and ShapeNetCore we set $M=64$ to reduce the processing time while obtaining approximately the same classification performance. Figure \ref{fig:param-tuning}(a) shows that increasing the complexity of our spatial function by setting a larger $M$ improves the performance of CompositeNets on challenging, real-world datasets such as ScanNet with a marginal increase in the number of parameters.

\noindent\textbf{Reducing the number of parameters.} In the second experiment, we train different configurations of our CompositeNets on the ModelNet40 dataset~\cite{WuModelnet}. In this case, we investigate how the OA varies when changing the output dimension of the spatial function, i.e. when setting $K\in\{16,12,8,4\}$. We compare the OA of our CompositeNets with the original KPConv~\cite{HuguesKPConv} when varying the number of centers $M\in\{16,12,8,4\}$. In fact, in KPConv, $M$ regulates the size of the weight tensor $\widetilde{W}$, which contains the majority of the layer parameters (see Figure \ref{fig:composite-vs-kpconv}(b)), thus playing a role similar to $K$ in our composite layers.

Figure \ref{fig:param-tuning}(b) shows the overall accuracy of each network and the overall number of parameters of the most and least complex versions of each network. On the one hand, KPConv achieves the best performance (when $M=16$), but its accuracy gets substantially worse when reducing the number of parameters by setting $M=8$ and $M=4$. On the other hand, both our convolutional and aggregate composite layers retain superior performance for low $K$. These results highlight the benefit provided by the design flexibility of our composite layers, which allows us to substantially reduce the number of parameters (--75\% passing from $K=16$ to $K=4$) without compromising the accuracy, which decreases by approximately 0.01. In fact, the smaller number $K$ of spatial function components is counterbalanced by a constant number of centers, which we set to $M=16$ in all the experiments. In contrast, the accuracy of KPConv is more sensitive to $M$ (--0.05 passing from $M=16$ to $M=4$), which is the only hyperparameter we can tune to reduce the number of parameters of a KPConv layer. Moreover, the reduction in the number of parameters is less substantial compared to our CompositeNets (--38\% passing from $M=16$ to $M=4$) since tuning $M$ does not change the number of parameters of the residual blocks.

\noindent\textbf{Combining convolutional and aggregate layers.} In this experiment, we implement three different versions of our CompositeNets using 3 convolutional composite layers (C) and 2 aggregate ones (A), in the following configurations: AACCC, CACAC, CCCAA. We train these networks for classification on ModelNet40~\cite{WuModelnet} and compare their number of parameters, OA, and AA with our convolutional (CCCCC) and aggregate (AAAAA) CompositeNets in Table \ref{tab:ablation}. 

These results show that the AACCC configuration outperforms our convolutional and aggregate CompositeNets in both OA and AA. This is probably due to the fact that aggregate layers enable achieving superior performance, as shown in our previous experiments, but also increase the number of parameters, which might lead to overfitting on small datasets such as ModelNet40~\cite{WuModelnet}. Using aggregate layers in the first stages of the network and convolutional layers in the final ones significantly improves the performance without excessively increasing the number of parameters.

\begin{table}[t]
\centering
\renewcommand{\arraystretch}{0.75}
\setlength{\tabcolsep}{12pt}
\caption{Overall (OA), Average Accuracy (AA) on ModelNet40 and number of parameters of the CompositeNets in our ablation study. The architecture with 2 aggregate and 3 convolutional layers (AACCC) has the best trade-off between performance and number of parameters.}\label{tab:ablation}
\resizebox{0.8\textwidth}{!}{  
\begin{tabular}{c|ccccc}
\toprule
\textbf{Architecture} & CCCCC & AACCC & CACAC & CCCAA & AAAAA\\
\midrule
\textbf{\# params} & $3.8\cdot10^6$ & $4.2\cdot10^6$& $7.4\cdot10^6$ & $13.3\cdot10^6$ & $15.2\cdot10^6$\\
\textbf{OA} & 0.913 & \textbf{0.916} & 0.907 & 0.909 & 0.911\\
\textbf{AA} & 0.871 & \textbf{0.889} & 0.882 & 0.876 & 0.879\\
\bottomrule
\end{tabular}
}
\end{table}

\subsection{Discussion and Limitations}
Our experiments on classification, segmentation, and anomaly detection tasks show that the performance of our convolutional CompositeNet is on par with ConvPoint, which shares the same purely sequential architecture and has a comparable number of parameters. The main advantage of our composite layer is the design flexibility, i.e. the fact that the composite layer exposes two meaningful hyperparameters, namely, the number of RBF centers $M$ and the number of components of the semantic function $K$, which can reduce/increase the complexity of the spatial and semantic components to control the overall number of parameters and the computational load. To demonstrate the benefits of such flexibility, we perform ablation studies (Section \ref{subsec:flexibility}) where we show that we can enrich the spatial function to improve performance without significantly increasing the number of parameters of our CompositeNets (see Figure \ref{fig:param-tuning}(a)). Moreover, we can shrink the semantic function to reduce the number of parameters of our CompositeNets without compromising their performance (see Figure \ref{fig:param-tuning}(b)). Needless to say, such a control on the overall number of parameters can be crucial in practical applications, where computational resources might be scarce and training sets are small, which increases the risk of overfitting. We have also shown that existing point-convolutional layers, such as KPConv and ConvPoint, do not allow such customization.

Our results suggest that aggregated composite layers are more powerful, at the cost of having more parameters than convolutional ones. Our ablation studies also show that architectures combining convolutional and aggregate layers can provide a better trade-off between performance and number of parameters. We believe that these promising results can be further improved by architectural enhancement, which is beyond the scope of this work.


We are among the first to address unsupervised deep anomaly detection on point clouds, adopting the self-supervised approach in~\cite{golan2018deep}. Here, we show CompositeNets favorably compares against other networks based on point-convolutional operators as well as shallow baselines based on hand-crafted descriptors and autoencoders. This anomaly-detection approach shows two main limitations, regardless of the underlying deep neural network. First, it enables only discriminating between normal and anomalous point clouds, and cannot be directly used to detect anomalous regions in 3D shapes. This is an emerging problem whose first benchmark has been recently released~\cite{bergmann2022mvtec3d}, and which we will address in future works. Second, most transformations used in~\cite{golan2018deep} for images cannot be employed on point clouds as these might not be well suited for the 3D domain or some classes of objects. For instance, all the self-supervised networks perform poorly on the class \emph{Lamp} from ShapeNet as this class contains both table and ceiling lamps, thus preventing the network from distinguishing rotations applied to lamps with different vertical orientations. In contrast, autoencoders~\cite{masuda2021toward} and shallow baselines seem to extract relevant features from these point clouds, achieving superior performance. 

\section{Conclusions and Future Work}\label{sec:conclusions}
We introduce the composite layer, a general formulation to define neural networks for point clouds, enabling more design flexibility  than existing point-convolutional layers such as KPConv and ConvPoint. Our experiments on both classification and semantic segmentation show that CompositeNets perform on par or better than ConvPoint, which has the same sequential architecture, and approach KPConv, which instead has a deeper,  residual architecture. Moreover, our results show that self-supervised anomaly detection networks perform substantially better than shallow baselines leveraging hand-crafted point cloud descriptors and existing anomaly-detection algorithms for point clouds based on variational autoencoders.

Future work will address the enhancement of our CompositeNets, leveraging deeper, residual architectures that comprise both convolutional and aggregate composite layers to improve performance, following the insights from our ablation studies. To address the intrinsic limitations of the self-supervised approach~\cite{golan2018deep}, we will investigate how to automatically learn suitable transformations directly from the point clouds of the training set as in \cite{qiu2021neural}, where such transformations have been successfully learned for signals and images. We will also design CompositeNets able to detect anomalous regions within point clouds, to be tested on real-world datasets such as the recent \emph{MVTec-3D-AD}~ \cite{bergmann2022mvtec3d}. Another research direction that combines the design flexibility of our composite layer and the self-supervised approach is unsupervised hyperparameter tuning and pre-training for 3D deep learning~\cite{zhang2021self}, which are paramount since annotated 3D datasets are not as large and widespread as 2D image datasets.

\section*{Acknowledgments}
This work is supported by the PNRR-PE-AI FAIR project funded by the NextGeneration EU program. We thank NVIDIA for granting four RTX A6000 GPUs to Politecnico di Milano with the Applied Research Accelerator Program.

\bibliography{bbl.bib}

\noindent\textbf{Alberto Floris} graduated in Computer Science and Engineering at Politecnico di Milano in 2021. He is currently a Data Scientist at Skyqraft AB, where he develops machine-learning algorithms for large-scale LiDAR point clouds. Part of this work was done during his internship at STMicroelectronics.

\noindent\textbf{Luca Frittoli} received the Ph.D. in Information Technology at Politecnico di Milano in 2022. He is currently a Data Scientist at lastminute.com, where he develops machine-learning algorithms for the online travel business. Part of this work was done during his Ph.D. at Politecnico di Milano.

\noindent\textbf{Diego Carrera} received the Ph.D. in Information Technology at Politecnico di Milano in 2018. In 2015 he was a visiting researcher at Tampere University of Technology. He is currently an Application Development Engineer at STMicroelectronics, where he develops quality inspection systems for wafer production. 

\noindent\textbf{Giacomo Boracchi} is an Associate Professor of Computer Engineering at Politecnico di Milano - DEIB, where he received the Ph.D. in Information Technology in 2008. Since 2015 he leads industrial research projects concerning outlier detection systems, X-ray systems, and automatic quality inspection systems. His research interests concern machine learning and image processing, in particular change/anomaly detection, domain adaptation, and image restoration.

\end{document}